# Automatic Identification of Alzheimer's Disease using Lexical Features extracted from Language Samples

M. Zakaria Kurdi

University of Lynchburg, VA

## Abstract

**Objective:** this study has a twofold goal. First, it aims to improve the understanding of the impact of Dementia of type Alzheimer's Disease (AD) on different aspects of the lexicon. Second, it aims to demonstrate that such aspects of the lexicon, when used as features of a machine learning classifier, can help achieve state-of-the-art performance in automatically identifying language samples produced by patients with AD.

**Methods:** data is derived from the ADDreSS challenge, which is a part of the DementiaBank corpus. The used dataset consists of transcripts of Cookie Theft picture descriptions, produced by 54 subjects in the training part and 24 subjects in the test part. The number of narrative samples is 108 in the training set and 48 in the test set. First, the impact of AD on 99 selected lexical features is studied using both the training and testing parts of the dataset. Then some machine learning experiments were conducted on the task of classifying transcribed speech samples with text samples that were produced by people with AD from those produced by normal subjects. Several experiments were conducted to compare the different areas of lexical complexity, identify the subset of features that help achieve optimal performance, and study the impact of the size of the input on the classification. To evaluate the generalization of the models built on narrative speech, two generalization tests were conducted using written data from two British authors, Iris Murdoch and Agatha Christie, and the transcription of some speeches by former President Ronald Reagan.

**Results:** using lexical features only, state-of-the-art classification, F1 and accuracies, of over 91% were achieved in categorizing language samples produced by individuals with AD from the ones produced by healthy control subjects. This confirms the substantial impact of AD on lexicon processing.

## 1. Introduction

Alzheimer's Disease (AD) is a type of dementia that is a progressive neurodegenerative disease that affects cognitive function, including language processing. One of the most significant language-related changes that occur in individuals with AD is a decline in their lexicon, which refers to the vocabulary and words they use to communicate [1]. The loss of lexicon can be particularly challenging, as it can make it difficult for individuals with dementia to express themselves clearly and to understand others. Early detection of dementia is critical for ensuring timely and appropriate treatment, as well as for improving patient outcomes [2]. In recent years, there has been growing interest in using machine learning with different types of linguistic features as a means of detecting AD at an early stage. Such an approach has several advantages as it is less intrusive, has virtually no side effects, and is cheaper than traditional approaches [3].

One promising approach is the use of lexical features, which are linguistic elements related to vocabulary and word usage. This study uses an automated analysis of transcribed speech as a screening tool for AD.

Although there has been a plethora of works that used lexical features of different types to detect AD and dementia, lexicon wasn't covered with enough depth and width. While current studies have shown promising results in using lexical features for detecting AD, there are still several gaps in our understanding of the impact of AD on lexicon processing. One major limitation of existing studies is that they often rely on a small number of lexical features, limiting the generalizability of their findings. Additionally, there is a lack of consistency in the types of lexical features that are measured and analyzed, making it difficult to compare results across studies.

To overcome these limitations, this study covers 99 lexical features. Some of these features have been used in previous studies about dementia, such as Brunet index, and Type Token Ratio (TTR). Other features have been used in other areas but have not been applied to dementia, such as sentiment analysis, Text Focus, and knowledge depth. Finally, some new features have been proposed in this paper, such as Point Wise Mutual Information (PMI) of word embeddings and diversity of Ngrams. To the best of our knowledge, no such extensive lexical level analyses have been carried out on bodies of linguistic production in AD. Using such many lexical features within the same study gives the advantage of comparing the benefits of those features individually and identifying the optimal combinations of those features in AD classification. Hence, this paper aims to combine two goals. First, study the impact of dementia on a wide number of lexical complexity measures. Second, show how these features, when grouped, can help detect dementia based on the written and transcribed spoken linguistic production of a subject. To achieve this second goal, an extensive evaluation is carried out to examine several key factors, such as the optimal set of features, the optimal input size, and the generalization of the trained models to detect dementia in different tasks.

This paper is organized as follows. In section 2, a survey of the relevant works about the impact of dementia on lexicon and dementia detection with machine learning is provided. The methodology and the used data set are presented in section 3. In section

4, a presentation of the lexical features adopted in this study along with their Anova stats. Machine learning experiments and their results are presented in section 5. The conclusion and future work will be presented in section 6.

## 2. Related Work

AD in its different types is affecting a large proportion of seniors around the world. Recent studies have shown that early diagnosis of dementia may help to delay its symptoms [2]. Clinical diagnosis of dementia is an intrusive procedure that is both expensive and stressful to patients. On the contrary, recent studies have shown the potential of machine learning and classification of linguistic features as a tool for detecting and monitoring cognitive decline in individuals with dementia.

Three types of work have been conducted about dementia and its impact on the linguistic abilities of the patients. First, some theoretical works have been performed. For example, Hier *et al.* [1] used a standardized picture description task of 39 patients with dementia to show that lexical deficits tended to be more severe than syntactic ones. Other research conducted longitudinal works on novels of Iris Murdoch, who was diagnosed with dementia after the publication of her last novel, concluded that lexical decline across her different novels, as the author becomes older, is more substantial than with syntax [4, 5]. In a follow-up of the works of [4-6], conducted a longitudinal analysis of the works of Iris Murdoch, , who was diagnosed with dementia after the publication of her last novel, Agatha Christie, an author who was suspected of having dementia during her last years, and the novelist P. D. James, who aged healthily. They used several measures of lexical complexity, such as Type Token Ratio, phrase repetition, and word-type introduction rate (WTIR) as well as knowledge from other linguistic levels such as syntax and disfluencies. They concluded that signs of dementia can be found in diachronic analyses of patients' writings including lexical aspects of texts.

Based on theoretical findings like the ones above, several works were conducted to build machine learning classifiers to diagnose patients with dementia based on their language production. Among the studies that investigated lexical features for dementia detection, the work by Fraser *et al.* [7] is worth mentioning. These authors examined the use of lexical richness measures, such as Moving-Average Type Token Ratio (MATTR), Type-Token Ratio (TTR), mean length of utterance, Brunet Index, and Honoré's statistic to distinguish between positive AD cases and healthy controls. The results showed that lexical richness measures were effective in differentiating between the two groups, with individuals with AD exhibiting lower lexical richness scores. Orimaye *et al.* [8] developed several machine learning models on 198 language transcripts from the DementiaBank[1]. Those transcripts were evenly selected from 99 patients with probable AD and 99 health control subjects. They combined syntactic features, such as average dependencies per sentence and average number of predicates, with lexical features, such as word count, word repetitions, and unique words, with the counts of specific Ngrams of words, like *the window*, *is open*, and *girl is*. They reported an AUC score of 0.93 on the 1000 Ngram model and 0.82 on the syntactic and lexical model. The main issue with this study is that it didn't rely on an extensive number of features that cover systematically the linguistic aspects that can be affected by AD. Another drawback, that was acknowledged by the authors, is that Ngrams of words are known to have scalability issues when applied to different tasks.

In 2020, several researchers from the University of Edinburg and Carnegie Mellon University launched a campaign about dementia called the ADReSS Challenge [9]. This challenge aimed to make available a benchmark dataset of spontaneous speech, which is balanced in terms of age and gender, to help compare different approaches of dementia detection. The ADReSS challenge consists of two tasks, the first of which is related to this work. This task is about AD classification task, where the participants must produce a model to predict whether a speech session is produced by someone with AD or a healthy subject. Several approaches have been proposed within the ADReSS Challenge campaign, some relied on speech data, transcribed speech, or both. Among those approaches, [10] used a two-stage hybrid architecture as a baseline. This architecture first uses VGGish is a pretrained Convolutional Neural Network from Google, for audio feature extraction, then they used Support Vector machine (SVM) and Neural Network NN to detect dementia. They also proposed DemCNN raw waveform-based convolutional neural network model that was 0.636 accurate, 7% more accurate than the baseline. According to the results sheet[2] published by the Challenge's organizers' this system achieved 0.875 on the campaign's evaluation test set. Yuan *et al* [11] proposed a model that fine-tuned BERT and ERNIE, pre-trained modern language models, to detect dementia. They achieved 0.896 accuracy on the test set of the ADReSS challenge, which is the leading performance of this campaign. Edwards *et al*. [12] adopted a combination of audio features and word and phoneme features. They analyzed the text data at both the word level and phoneme level. Experiments with larger neural language models did not result in improvement, given the small amount of text data available. However, they showed that the phoneme representation helps distinguish between positive and negative cases. This approach gave an F1 score of 0.854. Deep learning led to key unprecedented achievements within NLP. However, like other neural networks, this paradigm does not provide clear clues to understand the underlying problem. It is also known for demanding large amounts of data to achieve optimal performance, which is hard to find within the medical field.

Yamada *et al.* [13] collected speech responses in Japanese from 121 older adults comprising AD, Dementia with Lewy Bodies (DLB), and cognitively normal (CN) groups and investigated their acoustic jitter and shimmer features, prosodic features, such as pitch variation, pause duration, phoneme rate per second, and linguistic features, such as the number of correct answers and Type Token Ratio (TTR). They reported an F1 score of 0.864 for AD vs CN. Despite the promising results of this work, it doesn't offer either a systematic or a theoretically motivated approach.

---

[1] https://dementia.talkbank.org/

[2] https://dementia.talkbank.org/ADReSS-2020/2020Results.xlsx

The above-mentioned studies suggest that lexical features can be effective indicators of dementia, and that the use of natural language processing techniques can further improve the performance of the classifier to detect them. However, the previous studies lack systematic in-depth coverage of those features.

## 3. Methodology

### 3.1 Objective

The main aim of this paper is to explore systematically the discriminatory impact of lexical features in distinguishing between patients with AD and healthy users. Such exploration will help lay the ground to build a machine learning classifier that can achieve this task automatically. Hence, the conducted work is organized into two main phases. First, individual examination of the impact of AD on carefully selected lexical features that cover different aspects of lexicon, such as precision, diversity, and density. Second, finding the optimal set of features and the optimal length of input so that the classification accuracy and F1 are optimal.

### 3.2 Hypotheses

As shown in section 2, several previous studies suggested that dementia impacts lexicon more importantly than other key linguistic levels, such as syntax [1, 4, 5]. This leads us to assume that the lexicon of the subjects who are normally aging is less affected than the one of patients with AD. Furthermore, given the types of lexical features covered in previous works, it is assumed that AD impacts all aspects of lexicon that we are going to cover in this paper. Hence, AD is expected to cause a sharp decrease in density, diversity, sophistication, and specificity. The psychological and miscellaneous features are also expected to show some changes toward less complexity with patients suffering from AD. Since no previous systematic study of lexicon have been conducted, to the limit of our knowledge, it is hard to predict which of the lexical features will be impacted more by AD. Therefore, feature ranking is going to be one of the goals of this paper.

### 3.3 Datasets

The first dataset used in this study is the one provided by the ADDReSS challenge organizers [9]. This dataset is a part of the DementiaBank corpus, which is in its turn a part of the larger TalkBank project [14]. The dataset consists of recordings and transcripts of Cookie Theft picture descriptions, by 54 subjects in the training part and 24 subjects in the test part. The number of narrative samples is as follows: the training set contains 108 samples, and the test set contains 48 samples. Only the transcripts were used in this study. The ADDReSS dataset competition is adopted in this study because it is balanced in terms of age, gender, and conditions (see [9] for the distribution of genders, ages, and conditions of the participants). On the other hand, given that the ADDReSS data set was used in several previous studies, it makes it possible for us to compare our results with those reported previously on this data set.

Three other data sets were collected from two different renowned authors and a public personality, to test the system's generalization. Random passages of about 350 words were collected from those texts and speeches as described below.

Previous works, like [15, 16], suggested that President Reagan had early signs of Alzheimer's during his second presidency, years before his formal diagnosis in 1994. 40 passages were randomly selected from 10 of President Reagan's speeches. 20 passages were picked from speeches between 1964 and 1980, before he presumably started to show signs of dementia. The other 20 were selected from speeches made between 1989 and 1990, when he presumably started to show signs of dementia.

The second data set was collected from novels written by the British novelist Iris Murdoch, who died with Alzheimer's. Previous works suggested that the quality of her language was deteriorating with time [4, 6]. A corpus made of forty-four passages selected randomly from different parts of three of her novels: *Jackson's Dilemma*, published in 1995 that was written when she was suspected of having dementia, and two of her earliest novels: *Under the Net*, published in 1954, and *The Sandcastle*, published in 1957.

According to Le *et al.* [6], the British novelist Agatha Christie showed signs of dementia in her late writings. Hence, a data set is made of thirty-eight passages extracted from three of her novels: two of her earliest novels: *The Mysterious Affair At Styles, 1921* and *The Secret Adversary*, 1922 and her penultimate one: *Elephants Can Remember*, 1972, which was written when she was suspected of having dementia. The 19 passages extracted from her earlier works are considered as negative cases, while the 19 passages extracted from *Elephants Can Remember* are deemed as positive.

## 4. Feature Set

This section aims to describe the impact of dementia on individual features. For every feature, a brief description and justification will be provided along with its one-way Anova's F-score (hereafter Anova). Three other ranking methods will also be considered: reliefF, information gain, and $\chi^2$. Information about those three methods ranking will be provided when the feature is among the top features of these methods' ranked lists.

### 4.1 Measures of Lexical Density

Lexical Density (LD) refers to the percentage of content or lexical words in a given text or language sample. According to [17], LD is also a measure of the information density of a text. LD has been studied in relation to dementia. For example, [18] analyzed lexical-semantic and acoustic features of picture descriptions and found a decline in LD in texts produced by patients with AD. As seen in the density formula (figure 1.1), a higher LD score rate indicates a greater concentration of lexical or content words and a lower proportion of functional words in a text or a speech sample. Lexical words are words that are open, whose number is theoretically non-limited. The closed class words typically include grammatical words, whose number is limited in languages. Unfortunately, such a broad definition is not sufficient for algorithmic implementation, as there is a disagreement in the literature about the categories of words to include in each class. For instance, O'Loughlin [19] considers all adverbs of time, manner, and place as lexical. On the other hand, [20, 21] consider the following categories as lexical: nouns, adjectives, verbs (excluding modal verbs and auxiliary verbs), adverbs with an adjective base like *fast*, and those formed by adding the suffix *-ly* to their end, like *especially*. Hence, to account for the above divergence, two versions of LD were implemented. In the first version (LD1), the following word categories are considered as lexical: the verbs, including modals, the adverbs, including comparative and superlative adverbs, the adjectives, including comparative and superlative adjectives, gerund, present participle, nouns, and proper nouns. In the second version (LD2), implemented, the following categories of words are considered as lexical: all the common and proper nouns, adjectives, including comparative and superlative adjectives, as well as comparative and superlative adverbs. Verbs are considered as lexical except the modals *have* and *be*. Adverbs are considered as lexical when they end with the suffix *-ly* or when they have the same form as an adjective like *half*, *late*, or *low*. As shown in figure 2, the ANOVA stats of these two versions are not significant. A possible interpretation of this result is that the information density is the same between the populations with and without dementia, given that the task is the same.

Word Frequency Index (WFI) is a variant of LD where the sum of the functional words is divided by the total number of words (figure 2.1). As shown in figure 2, WFI did not yield a significant Anova stat either. This result is probably because the task of the conversation is limited.

| # | Measure | Equation | |
|---|---------|----------|--|
| 1 | Lexical Density | $LD = \dfrac{N_{Lex}}{N}$ | $N_{lex}$ is the number of lexical words<br>$N$ is the total number of words |
| 2 | Word Frequency Index (WFI) | $WFI = \dfrac{\sum_{i=1}^{N} count(w_i)}{N}$ | $w_i$ is a functional word from the text<br>$N$ is the total number of words |

**Figure 1.** Lexical Density equations

One can look at density from the opposite angle too. Therefore, some used Functional Word Ratio (FWR), which is the ratio of the number of function words, such as articles, pronouns, and prepositions, to the total number of words in a text or speech sample. A higher FWR indicates lower lexical density. FWR has been used in the studies of language development and author profiling [22, 23]. FWR did not yield a significant Anova stat (figure 2). Despite that, FWR is among the top 25 features within the ranked lists of reliefF and information gain.

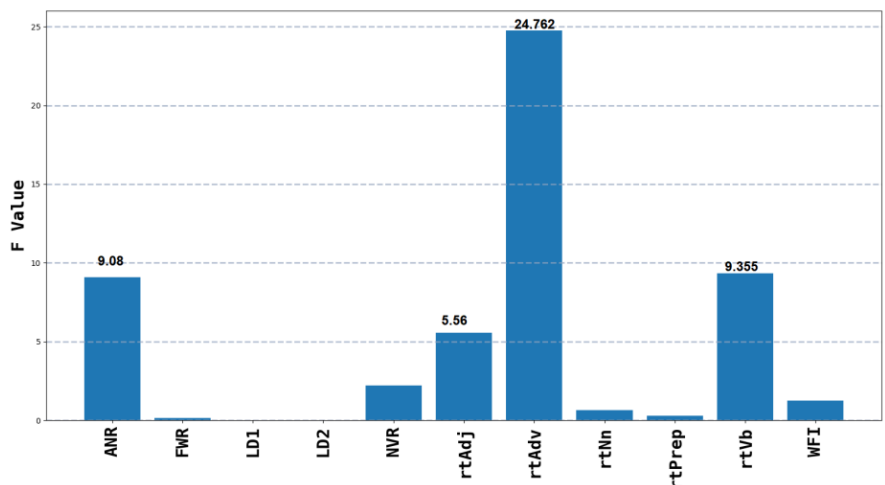

**Figure 2**. ANOVA ranking of the features of lexical density, the features ANR, rtAdj, rtAdv, and rtVB are significant with $p \leq 0.03$

Another approach to measuring density is by taking every word's morphosyntactic category or its Part Of Speech (POS) tag and calculating its ratio to the number of all words within the text. Five features of this type are considered: the ratio of adjectives (rtAdj), the ratio of the nouns (rtNn), the ratio of the verbs (rtVb), the ratio of prepositions (rtPrep), and the ratio of adverbs (rtAdv). Some more specific ways to consider the ratio of morphological categories have been explored in the literature, by taking the ratio of a given category to another, like the Noun-to-Verb Ratio (NVR). A higher NVR indicates higher lexical density. Adjective-to-

Noun Ratio (ANR) is the ratio of adjectives to nouns in a text or speech sample. A higher ANR indicates lower lexical density. This aspect of density was considered by Ahmed *et al.* [24] in their study of 12 proportional frequencies of open class (nouns, verbs, and descriptive terms) and closed class or grammatical words. They found no significant overall differences, after Group comparisons between normal controls and patients with AD. On the other hand, using noun and verb naming, sentence completion, and narrative tasks, Kim and Thompson [25] examined the nature of verb deficits in 14 individuals with Probable Alzheimer's Disease (PrAD). Production was tested, controlling both semantic and syntactic features of verbs. This study also covered noun and verb comprehension and a grammaticality judgment task. Results showed that both PrAD subjects showed impaired verb naming and that NVR with PrAD is consistent with grammatical aphasia. As shown in figure 2, only the features ANR, rtAdj, rtAdv, and rtVB have significant Anova stats. Furthermore, the features ANR, FWR, rtAdj, rtAdv, rtVB, and NVR, are among the top 25 features within the relief ranked list of features. The features rtAdj, rtAdv, rtVB, rtPrep, and NVR are among the top 25 features $\chi^2$ ranked list of features. The feature rtAdj, rtNn, rtVB, and FWR are among the top 25 features within the information gain ranked list. Overall, these results show the importance of most of the lexical density used in this paper play are strong indicators of AD.

### 4.2 Lexical Diversity

Lexical Diversity (LD) or lexical variation is a way to measure the variety of words or vocabulary used to express ideas about a given subject. LD is an important aspect of language that can impact the effectiveness and clarity of written and spoken communication. Several measures of lexical diversity have been studied within the context of works about language acquisition and education, as well as about dementia [26, 5, 27].

The intuitive way to describe LD is the count of the different lexical forms in the text, or the size of the vocabulary. This is called the Number of Different Words (NDW). NDW was used in areas like language acquisition [28] or ESL [27]. NDW was also used by [29] to identify the *core vocabulary* for adults with complex communication needs. The obvious limitation of NDW is its bias introduced by the text length. The longer the text, the bigger is the chance of observing different lexical forms. This limitation motivated the creation of several extensions to measure diversity independently of the text length. As seen in figure 3, the Anova stat of NDW is not significant.

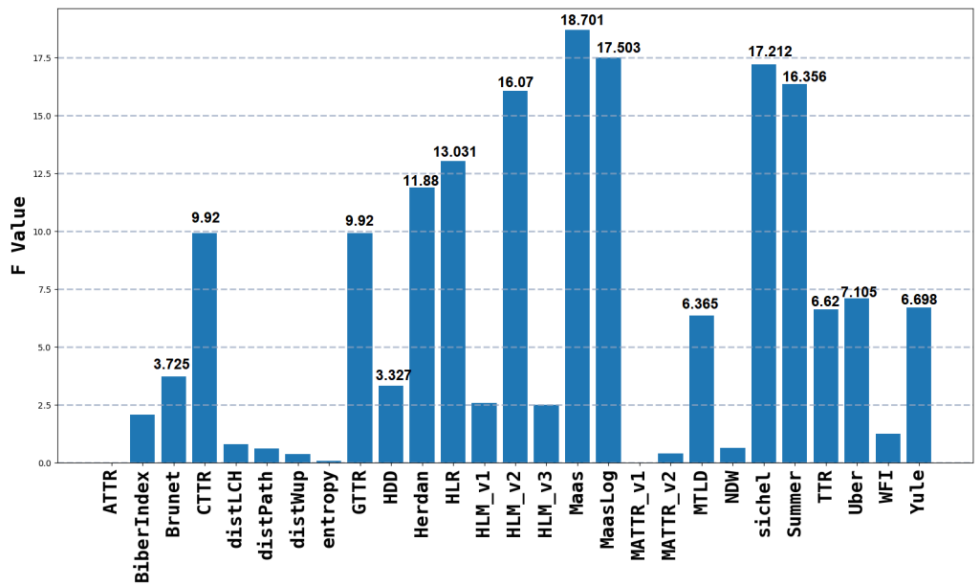

**Figure 3.** F-values of the Anova test, the following features are significant with P ≤ 0.05 Brunet, CTTR, GTTR, HDD, Herdan, HLM_v2, Maas, MaasLog, MTLD, Sichel, Summer Uber, and, Yule

Type Token Ratio (TTR) is a linguistic measure used to determine the lexical diversity of a text or a speech sample [30]. TTR is an extension of NDW, aiming to take into consideration the length of the text or speech sample. It is calculated by dividing NDW in a text by the total number of words in that text (figure 4.1). A high TTR indicates that a text has a greater variety of vocabulary and is more lexically diverse. TTR has been used in several studies about dementia with contradictory results. For example, TTR didn't give significant results in the study conducted by Shinkawa and Yamada [31] that aimed to characterize the atypical repetition of words on different days, which is observed in patients with dementia. While Le *et al.* [6] reported significant results in their longitudinal study about lexical and syntactic changes of three British novelists by comparing their works before and after they got dementia. These problematic results of TTR are probably due to its bias toward the size of the text, as was shown by [32, 33, 34], (see [26] for a detailed discussion), since the ratio decreases as the number of words in the text increases. As seen in figure 3, the Anova's stat of TTR is significant.

| # | Measure | Equation | |
|---|---------|----------|---|
| 1 | Type Token Ratio | $TTR = \dfrac{NDW}{N} * 100$ | N is the number of words |
| 2 | Guiraud Index | $GTTR = \dfrac{NDW}{\sqrt{2N}}\,100 = TTR\sqrt{\dfrac{N}{2}}$ | NDW is also the size of the vocabulary V |
| 3 | Caroll Index | $CTTR = \dfrac{NDW}{\sqrt{N}}\,100 = TTR\sqrt{N}$ | |
| 4 | MATTR | $MATTR = \dfrac{\sum_{i=1}^{W} TTR_i}{W}$ | W is the number of windows |
| 5 | Herdan Index | $C = \dfrac{\log(V)}{\log(N)}$ | V (or Vocabulary) is the number of different words |
| 6 | Brunet Index | $BI = N^{V(-.165)}$ | |
| 7 | Summer Index | $SI = \dfrac{\log(\log(V))}{\log(\log(N))}$ | |
| 8 | Uber Index | $Uber = \dfrac{\log(N)^2}{\log(N) - \log(V)}$ | |
| 9 | Yule Index | $Yule's\ k = 10^4 \dfrac{\{\sum_{r=1}^{N} V_r r^2\} - N}{N^2}$ | $v_r$ is the number of word types, which occur $r$ times in a text of length $N$ |
| 10 | Sichel Measure | $Sichel = (1 - \dfrac{N}{V \log(N)})\,100$ | |
| 11 | Maas index | $Maas = \dfrac{\log(N) - \log(V)}{\log^2(N)}$ | |
| 12 | MaasLogIndex | $MaasLog = \dfrac{log(V)}{\sqrt{1 - \dfrac{log(V)}{log(N)}}}$ | |
| 13 | Honoré Lexical Measure v1 | $HLM\_V1 = \dfrac{N}{\sqrt{\dfrac{N_{SingOc}}{V}}}$ | $N_{SingOc}$ = number of words with a single occurrence in the text |
| 14 | Honoré Lexical Measure v2 | $HLM\_V2 = 100\,\dfrac{\log(N)}{1 - \left(\dfrac{V}{N}\right)}$ | V = total number of different words in the text (i.e., the vocabulary) |
| 15 | Honoré Lexical Measure v3 | $HLM\_V3 = \dfrac{N}{\sqrt{V}}$ | |
| 16 | Entropy | $H(x) = -\sum_{x \in X} p(x)\,\log_2(p(x))$ | $p(x)$ is the probability of a word $w$ occurring in the text $T$. |

**Figure 4.** Equations of the Different Diversity Features

To compensate for the change in the size of the text and consequently turn TTR into a constant over the whole text, several mathematical transformations of TTR were attempted. For instance, Guiraud's corrected TTR (GTTR) [35] is one of these transformations (figure 4.2). Caroll [36] also proposed a similar transformation called Caroll Type Token Ratio (CTTR), (figure 4.3). Both CTTR and GTTR yielded a significant Anova stat and are among the top 25 features within the $\chi^2$ ranked list.

A more recent transformation called moving-average type–token ratio (MATTR) has been proposed [37]. MATTR addresses LD using a moving window that calculates the TTRs for each successive window of a given length. The final MATTR score is calculated as the average of the TTRs of the windows (figure 4.4). Two versions of MATTR have been implemented. In the first version (MATTR_v1), a fixed window of ten words with a gradient move of five words was used. In the second version (MATTR_v2), a window of 1/10th the text size and moves with half window size. Neither version yielded a significant Anova stat, however MATTR_v1 ranks among the top 50 features of the information gain list.

The Measure of Textual Lexical Diversity (MTLD) is another alternative to TTR. MTLD is calculated as the mean length of sequential word strings in a text with a given TTR value [38]. As shown in figure 3, the one-way ANOVA stat of this feature are significant.

The $D$ measure is another approach to calculating the lexical diversity independently of the length of the text [39]. The $D$ measure is based on the predicted decrease of the TTR, as the size of the text increases. This mathematical curve is compared with actual data from the MJ and the Lancaster–Oslo–Bergen corpora (see [38, 40] for more details). Besides, McCarthy and Jarvis showed that *Vocd-D* is a complex approximation of the hypergeometric distribution, and to show this, they proposed an index that they called HD-D or HDD. The hypergeometric distribution being the probability of drawing a certain number of tokens of a specific type from a text sample of a certain size. As shown in figure 3, this HDD feature has significant ANOVA stat.

Dugast's or Uber Index (Uber) is a transformation of TTR designed to measure the lexicon diversity based on the frequency of different word lengths in a text [41], (figure 4.8). U was used by Lissón and Ballier [42] to study vocabulary progression in foreign language learners and by Nasseri [43] to model the lexical complexity of academic writing. As shown in figure 3, Uber yielded a significant Anova stat.

Yule proposed a measure $K$ of repetition to account for lexical diversity within the context of his work on author identification, with the assumption that $k$ would differ between authors (figure 4.9) [44]. Yule has a significant Anova stat.

The Brunet Index (BI), also known as $W$, is a measure of the lexical diversity of a text or speech sample. BI is calculated based on the number of unique words within a text sample (figure 4.6). A higher BI indicates higher lexical diversity. BI was used as a feature in several dementia classification models like [45, 7, 46, 15, 47]. As shown in figure 3, BI produced a significant Anova stat.

Honoré's Lexical Measure (HLM) is another alternative to TTR to measure lexical diversity of a text or speech sample [48]. The main assumption behind HLM is that the frequency of a word is inversely proportional to its rank in the frequency list. HLM was used as a feature to detect dementia with machine learning in several previous studies like [45, 7, 15]. Based on the literature, three versions of HLM were implemented (figure 4.13, 4.14, and 4.15). As shown in figure 3, only the second version of HLM has a significant Anova stat.

Summer Index (SI) is a measure of lexical diversity (figure 4.7), that was used by Lissón and Ballier [42] to study vocabulary progression in second language learning. As shown in figure 3, SI gave a significant Anova stat.

Herdan's Index (HI) is another lexical diversity measure that is also known as LogTTR [49] (figure 4.5). HI has been used in studies about foreign language progression [42]. As shown in figure 3, HI yielded a significant Anova stat.

Maas Index (MI) is a different measure of diversity proposed by Maas [50] to improve TTR. MI is based on the association between the number of unique words and the total number of words (figure 4.11). A variant of this measure called Maas log (MaasLog) has also been used in literature (figure 4.12) [42]. Both measures yielded a significant Anova stat (figure 3).

Sichel was used to study language progression by foreign language learners [51] and it was also used as a feature in the model proposed by Wang *et al.* [15] to detect AD (figure 4.10). As seen in figure 3, SM yielded a significant Anova stat.

Entropy is a thermodynamic concept that was proposed to measure the degree of disorder of randomness (figure 4.16). Entropy has been used to measure the uniformity of the vocabulary distribution [52]. Hernández-Domínguez *et al.* [46] showed that Entropy has significant but weak correlation with cognitive impairment. However, as shown in figure 3, entropy did not yield a significant stat, but it figures among the top 50 features of reliefF's ranked list.

### 4.3 Diversity through Lexical Focus and Point Wise Mutual Information (PMI)

Another way to measure the diversity of a text is by measuring the focus of the text. Text focus was proposed by [53] as the mean distance between keywords extracted from the text. Key words are extracted using the Rapid Automatic Keyword Extraction (RAKE) algorithm [54]. Three distances between the keywords are then calculated using the wordnet dictionary: Wu-Palmer (distWUP) Similarity[3], Path similarity[4] (distPath), the Leacock-Chodorow (distLCH) Similarity[5]. The features *distWUP* and *distPath* do not have a significant Anova stat but *distPath* is among the top 25 ReliefF's ranked features. Even though the Anova stat of *distLCH* is not significant, this feature is among the top 25 features in the ranked lists of reliefF, $\chi^2$, and information gain.

---

[3] based on the depth of the two senses in the hierarchy and that of their least common subsumer.
[4] returns a score based on the shortest path that connects the senses in the hypernym hierarchy.
[5] based on the shortest path that connects the senses and the maximum depth of the hierarchy in which the senses occur.

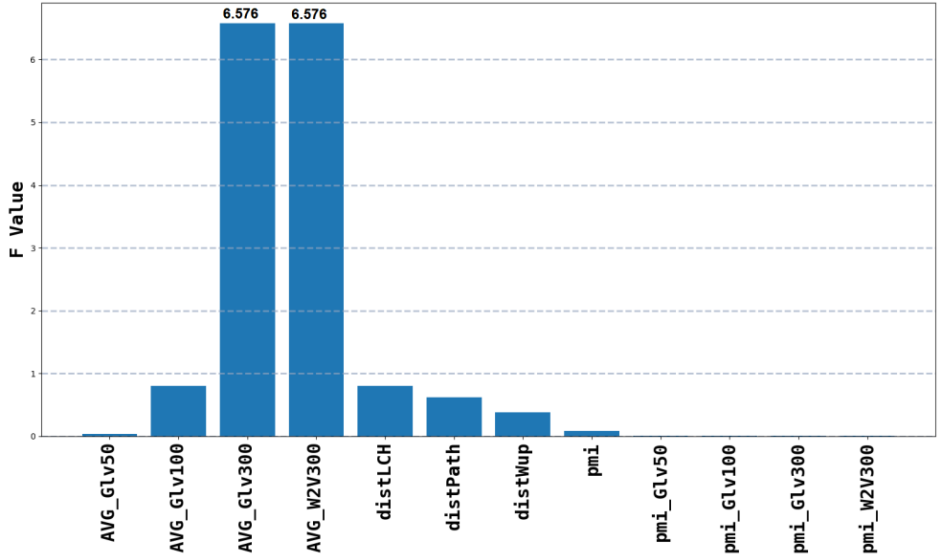

**Figure 5** F-values of the Anova stats of the text focus and PMI features, features AVG_Glv300 and AVG_W2V300 are significant with p=0.01

A fourth approach to measure the text focus is also considered. This approach consists of using the average of the cosine distances between the glove and word2vec vectors of the keywords respectively (figure 6.1). Given the possible impact of the size of those vectors, the experiments were conducted with the following sizes: 50, 100, 200, and 300. We got four features glove: AVG_GlvSIZE and two AVG_W2VSIZE (to avoid redundancy). Only the features: AVG_Glv100 and AVG_Glv300 and AVG_W2V300 have significant Anova stats. However, AVG_Glv50 is among the top 25 of ReliefF and AVG_Glv100 is among the top 25 information gain features.

1. Text Focus

$$focus(T) = \frac{\sum_{i=1}^{k-1} \cos(V_{w_i}, V_{w_{i+1}})}{k} \quad \forall w_i \in T$$

*T is the text sample*
*Cos is the cosine distance*
*$V_{wi}$ is the vector word i*
*K is the number of tokens in T*

2. PMI words

$$PMI(w_x, w_y) = \log 2 \frac{p(w_x, w_y)}{p(w_x)p(w_y)}$$

3. PMI word vectors

$$PMI(w_x, w_y) = \log 2 \frac{\cos(V_x, V_y)}{p(w_x)p(w_y)}$$

Figure 6. Equations of PMI and focus

In statistics, PMI is used to measure the association between two events. When it comes to lexicon, it indicates a high probability of co-occurrence, which is among others, a measure of semantic relatedness. Hence, PMI can be used as a measure of diversity. Two approaches to calculate PMI are adopted in this paper. First, PMI is calculated according to the traditional approach that consists of calculating the probability of co-occurrence of two terms within the same text (figure 6.2). The average PMI of the pairs of content words within the text is calculated. This approach did not give a significant Anova stat. However, it ranked among the top 25 features within the $\chi^2$, reliefF, and information gain. The second approach to PMI is an extension proposed in this paper to consider the semantic relatedness of the pairs of words (figure 6.3). Two types of word embedding have been used: word2vec and glove, with varying vector sizes: 50, 100, and 300 (pmi_glvSize) and one size for word2vec, to avoid redundancy: pmi_W2V300. All the PMI features of word embedding gave non-significant Anova stats. However, pmi_Glv50 is among the top 25 features of reliefF and pmi_Glv100, pmi_Glv300, and pmi_W2V300 are among the top 50 features of the same list.

### 4.4 Diversity of Ngrams of Words

Using the DementiaBank language dataset, Orimaye *et al.* [8] compared the 20 most frequent Ngrams between patients with preliminary AD and Healthy Elderly People and saw significant differences between those two groups. Such an approach relies more on pattern matching, that dependent on the literal aspect of the text. Therefore, this approach does not scale well to a change of the task. Hence, in this paper, three lexical diversity measures were extended to Ngrams: TTR, CTR, and GTTR. Every one of these measures was implemented with bi-grams, trigrams, and fourgrams. This gave us the nine measures presented in figure 7.

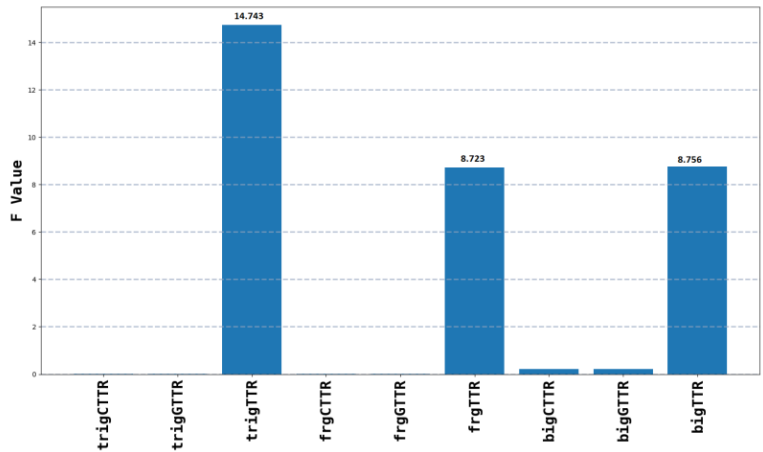

**Figure 7.** Ngram diversity measures. Only bigTTR, trigTTR, and frgTTR are significant with $p \leq 0.004$

Among those measures, only the three Ngrams of TTR gave significant Anova stats (figure 7). TrigTTR is also among the top 25 features in the information gain's ranked list. The features trigCTTR and trigGTTR are among the top 50 features of the information gain and $\chi^2$ ranked lists. The feature frgGTTR are among the top 50 features within the $\chi^2$ ranked list.

### 4.5 Lexical Sophistication

Lexical sophistication (Soph) is also called lexical rareness [55, 21] or Basic Lexical Sophistication. It is calculated as the ratio of the number of sophisticated words to the number of lexical words (figure 7.1). Hyltenstam considers as sophisticated the words whose frequency is larger than 7000 in the list of most frequent Swedish words [56]. Astell and Harley induced tip-of-the-tongue (TOT) states in their experiments with elderly participants, some of whom have PrAD [57]. The results of these experiments showed that TOT states occurred more often with patients with PrAD when the targets were words of low frequency and imageability. Given the above consideration, three versions of the Lexical sophistication are implemented. Sophistication (soph) is a basic version of lexical sophistication, where a word is considered sophisticated if its frequency rank is more than 3000. Word frequencies are obtained from the Word Frequency Data (WFD)[6], a freely available list of the 5000 most frequent words in English that is calculated based on the Contemporary American English Corpus [58]. The words that are not in the WFD are given the same frequency value, which is lower than the lowest score in the list. Words are stemmed using the Snowball Stemmer. This helps match the plural of the regular forms (e.g. *book*, *books*). The ANOVA stat of *Soph* is significant (figure 9).

| # | Measure | Equation | |
|---|---------|----------|---|
| 1 | LEXsph | $Lex_{soph} = \dfrac{N_{soph}}{N}$ | $N_{soph}$ is the number of sophisticated words<br>$N$ is the total number of tokens (that remain after filtering the stop words) |
| 2 | CLS | $CLS = \dfrac{\sum_{i=1}^{k} \text{WFD}(w_i)}{N}$ | WFD($w_i$) is the frequency of the word $w_i$ within WFD. |
| 3 | VSM | $VSM = \dfrac{V_{soph}}{V}$ | $V_{soph}$ is the number of sophisticated verbs<br>$V$ is the total number of verbs |
| 4 | VSM sq | $VSM\ sq. = \dfrac{V_{soph}}{\sqrt{2*V}}$ | |
| 5 | HLR | $HLR = \dfrac{N_{sin glOc}}{N}$ | $N_{singlOc}$ is the number of word with a singl occurrence in he text |

**Figure 8.** Equations of the features of sophistication

A second variant of lexical sophistication, the Continuous Lexical Sophistication (CLS), is proposed by [27]. It is about calculating the average frequency of the content words within a text, after removing stop words (figure 8.2). Like with *Soph*, words are also stemmed here. This feature did not give a significant Anova stat, but it is among the top ranked features by information gain and relief methods.

---
[6] http://www.wordfrequency.info/free.asp

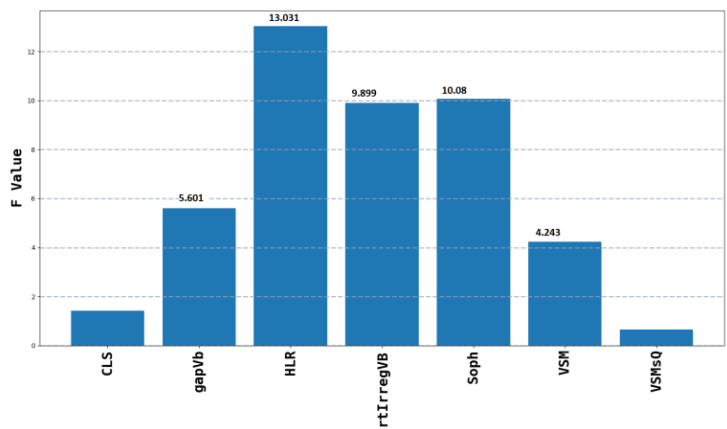

**Figure 9** Anova stats of the sophistication features, HLR, gapVb and VSM are significant, with p≤0.04

A related measure was proposed by Harley and King [59]: the verb sophistication measure (VSM). It is calculated as the ratio of the number of sophisticated verbs to the total number of verbs (figure 8.3). Sophisticated verbs are defined as the verbs that are outside of the list of the most frequent verbs. Harley and King used two lists with 20 and 200 verbs respectively in two different studies. In both studies, they reported a significant difference between native and non-native writers. In this paper, the McMillan English Dictionary was used. This dictionary provides the list of the 330 most frequent verbs[7] in English. To find the uninflected form of a verb, the verb conjugation module, provided within the Pattern.en toolbox[8], is used. As seen in figure 9, this feature yielded a significant Anova stat. It is also among the top 25 reliefF features. A modified version of VSM, that uses the square to reduce the sample size effect, was proposed by Wolfe-Quintero *et al.* [60] (figure 8.4). This version did not give a significant Anova stat, but it is among the top 50 $\chi^2$ ranked features.

GAP verbs (gapVB), sometimes called general all-purpose verbs or light verbs, are a limited set of high frequency, often mono syllabic, verbs with general semantic meanings, such as the verbs: *make*, *do*, and *go* [61]. As suggested by Maouene *et al.*, heavy verbs have a strong association with specific objects [62]. There have been conflicting findings concerning high usage of gapVB in children with Specific Language Impairment (SLI), but it is now thought that making extensive usage of gap verbs is a normal phase in both children developing normally and those with SLI [63]. Given the relevance of gap verbs to the process of language acquisition and verb specificity, they seem a relevant indicator of dementia. Hence, the ratio of gap verbs to the total number of verbs gave relevant Anova stats, it is also among the top 25 features in the three other feature selection lists: reliefF, $\chi^2$, and information gain. This result suggests that patients with dementia tend to use more generic gap verbs than control subjects.

Another measure related to verbs is the ratio of regular vs irregular verbs (rtIrregVB). Previous works didn't provide clear evidence of the difference in terms of lexical access to regular vs irregular verbs. For example, Feldman *et al.* [64] concluded that normal native speakers didn't show substantial difference in access to irregular verbs compared to the regular ones, while Justus *et al.* observed a dissociation between the two types of verbs [65]. Nonetheless, the divergence between the two types of verbs requires extra cognitive processing that can be altered by dementia. This assumption is confirmed by the significance of the Anova stats and by the fact that this feature is among the top 25 features of the three other considered feature selection methods: reliefF, $\chi^2$, and information gain.

Hapax Legomena Ratio (HLR) takes a local approach to sophistication, since it defines complexity as the ratio of the number of words that occur only once in a text or a speech sample to the total number of words in the sample. Hernández-Domínguez *et al.* [46] showed that HLR has a negative but significant correlation with cognitive impairment. HLR yielded a significant Anova stat and it is among the top 50 $\chi^2$ ranked features.

### 4.6 Lexical Specificity

Lexical Specificity (LSP) is about how precise the lexicon is used in a text or a speech sample. Lexicon is said to be precise when it designates a precise object or action. Vagueness and ambiguity are the result of using a lexicon that lacks specificity. LSP has been linked to dementia in the literature. For example, Eyigoz *et al.* showed that using the semantically generic terms, such as *boy*, *girl*, and *woman* instead of the more specific respectively: *son*, *brother*, *sister*, *daughter*, and *mother* to refer to the subjects in the picture is associated with higher risk of AD [66]. Le *et al.*, in their longitudinal study of the writings of three British novelists, showed that there is a significant decrease in specificity associated with dementia [6].

---

[7] http://www.acme2k.co.uk/acme/3star%20verbs.htm

[8] http://www.clips.ua.ac.be/pages/pattern-en

Several features have been proposed in this paper to capture LSP. First is the depth of knowledge required by a set of words. A word is considered to require more knowledge, or to be specialized, if it belongs to the list of University Word Lexicon (UWL) compiled by Xue and Nation [67]. The UWL is made up of 836 words. This feature gave significant Anova stat.

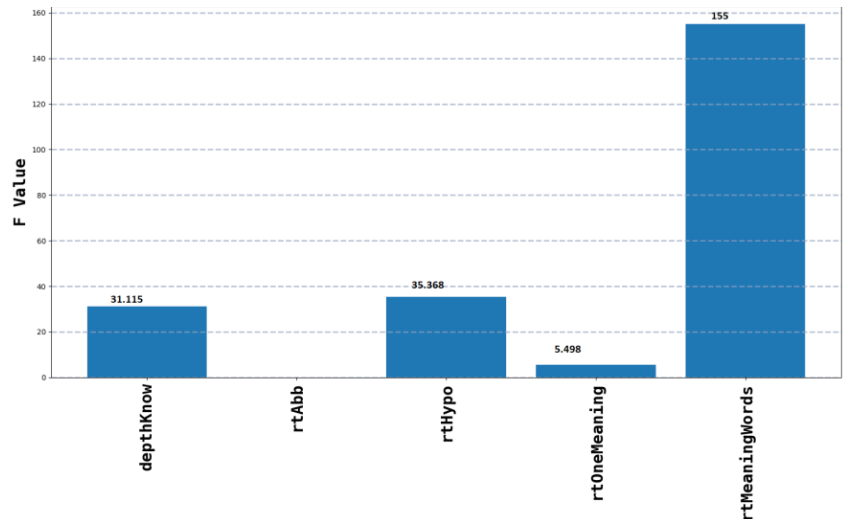

**Figure 10** Anova stats of specificity features, Significant features depthKnow, rtHypo, rtOneMeaning, rtMeaningWords, with $p \leq 0.02$

To capture the polysemy of the content words within the texts, those words that remain after removing the stop words, two approaches have been adopted. The first consists of calculating the ratio of monosemic words to the number of content words (rtOneMeaning) and the second consists of calculating the ratio of the total number of meanings to the number of words within the text (rtMeaningWords). WordNet is used to calculate both ratios. As shown in figure 10, both *rtOneMeaning* and *rtMeaningWords* yielded a significant Anova stats. The feature *rtOneMeaning* is among the top 25 $\chi^2$ ranked list and among the top 50 of the ranked list of information gain. The feature *rtMeaningWords* is among the top 50 features within the reliefF ranked list.

Abbreviations, such as *DIY*, *BTW*, and *LOL*, can be seen as a special case of monosemic words. Therefore, their ratio is calculated. This feature did not yield significant results, because of the rareness of usage of such type within the productions of patients.

Hyponyms are a way to capture the specificity of a word. The more hyponyms a word has the less specific it is. Using WordNet, the ratio of hyponyms to the words within the text (rtHypo) is calculated. This feature yielded a significant Anova stats. It is also among the top 25 of the three other lists of feature selection: reliefF, $\chi^2$, and information gain.

### 4.7 Psycholinguistic Aspects of the Lexicon

Several psycholinguistic aspects of the lexicon have been studied in the literature. These studies led to standard measures in psycholinguistics. Some of those key standards have been coded within the MRC psycholinguistic database[9], which covers 150837 words and provides 26 linguistic and psycholinguistic properties. In this paper, the following psycholinguistic features are considered as relevant to this study: Kucera-Francis Written Frequency (kf_freq), Kucera-Francis number of categories (kf_ncat), Kucera-Francis number of samples (kf_nsamp) [68], Thorndike-Lorge written frequency (TL_freq), Brown verbal frequency, Familiarity rating (familiarity), Concreteness rating, Imageability rating, Meaningfulness with Colorado Norms (Colorado) [69], Meaningfulness with Paivio Norms (Paivio), as well as the Age of Acquisition rating (ageAquis). Some of these features have been used in studies about dementia. For example, in their longitudinal study of three patients with semantic dementia, Le *et al.*, confirmed a significant impact of word imageability of spoken production of patients with dementia [6], which was shown before by Astell and Harley [57]. On the other hand, Bird *et al.* designed an experiment where they divided regular and exception words, where pronunciation did not follow directly from the orthography, based on their Kucera Francis frequency, and tested whether the patient could tell those words from false font strings, sequences of non-orthographic symbols [70]. The results of their study were not significant with both high and low frequency words.

These psycholinguistic features are calculated as follows. After extracting the list of the nouns from the text and their total number of occurrences *k* is counted. The scores of every psychological feature are extracted from the MRC dictionary and their average within the text is calculated. Finally, the ratio of the number of nouns that are in the MRC to the number of nouns in the text (rtNouns) is also used as a feature.

---

[9] http://websites.psychology.uwa.edu.au/school/MRCDatabase/uwa_mrc.htm

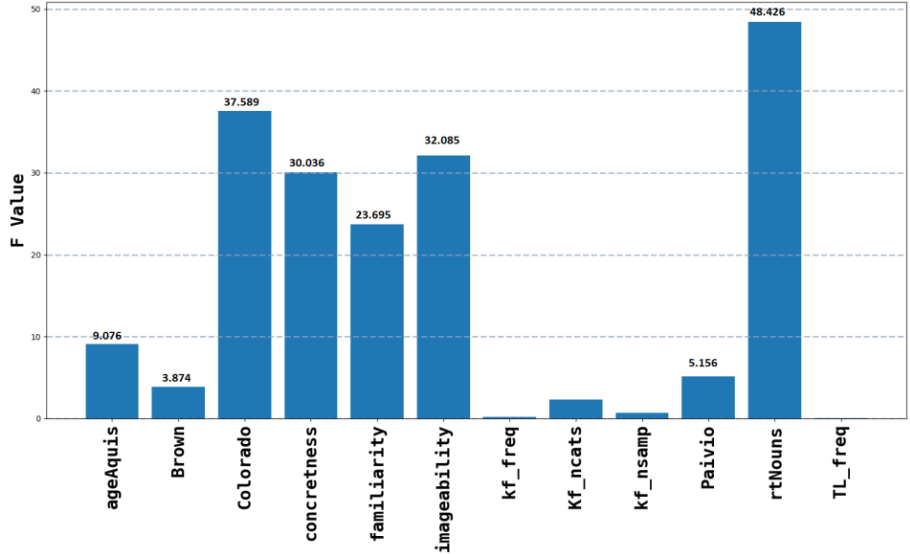

**Figure 11** Anova F of the psycholinguistic features of lexicon the significant features are: ageAquis, Brown, Colorado, concreteness, familiarity, imageability, Paivio, and rtNouns, with $p \leq 0.05$

As seen in figure 11, all the psycholinguistic features yielded a significant Anova stats except for the three Kucera Francis features and Thorndike-Lorge written frequency. Furthermore, several psycholinguistic features ranked high with the other feature selection approaches (table 1).

| Feature | Info. Gain | ReliefF | $\chi^2$ | Feature | Info. Gain | ReliefF | $\chi^2$ |
|---|---|---|---|---|---|---|---|
| ageAquis | - | - | x | Kf_freq | - | x | - |
| Brown | x | - | x | Kf_ncats | - | - | - |
| Colorado | x | x | x | Kf_nsamp | - | x | - |
| Concreteness | x | x | x | Paivio | x | - | x |
| imageability | x | x | x | rtNouns | x | x | x |
| familiarity | x | x | x | TL_freq | - | x | - |

**Table 1** Rankings of the psycholinguistics within the top 25 of information gain, reliefF, and $\chi^2$

As shown in table 1, Kf_ncats is the only feature that did not rank in any top list in addition to being not significant in Anova. This is probably because this feature is designed more for written language. On the other hand, the high rankings of many psycholinguistic features confirm their importance for dementia detection.

### 4.8 Miscellaneous Lexical Features

Discourse connectors, or simply connectors, are a simple and reliable way to measure the cohesion of a text or speech sample. Connectors, such as *and*, *so*, and, *also* indicate a complex content. Edwards *et al.*, used the number of coordinating conjunctions, and the number of subordinated conjunctions as features in their dementia classification system [12]. The reports in the literature don't provide a clear account of the impact of dementia on the use of discourse markers. For example, Davis and MacLagan [71] reported that the patient they followed in their longitudinal study didn't show changes in her use of *Uh*, the discourse marker that they focused on. On the other hand, in their study of poly-synthetic agglutinating language, [72] reported a decline in subordination with clausal connectors. In this study, a distinction is made between general connectors and argumentative discourse connectors, such as *but*, *therefore*, and *hence*, are a subset of discourse connectors that indicate a higher level of reasoning and argumentation, making them more complex. Two features of this type are calculated: the ratio of discourse connectors to the number of words and to the number of sentences, respectively (discWrd) and (discSent). Two similar features regarding the argumentative discourse connectors are calculated too: (discArgWrd) and (discArgSent).

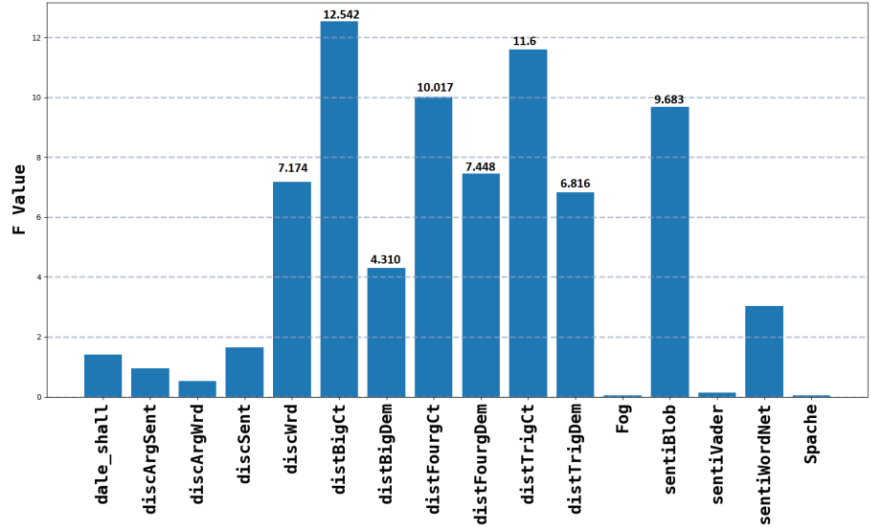

**Figure 12**. Anova F-scores of the miscellaneous features. Only discWrd and sentiBlob are significant with p ≤ 0.04

As seen in figure 12, among the four discourse features, only the feature discWrd yielded a significant Anova F-Score. This is due to the limited usage of argumentative discourse connectors within the cookie description task and to the difficulty of delimiting sentences within spoken language.

People with dementia tend to have issues with emotion regulation processes [73]. Therefore, it seems relevant to consider emotional lexicon as a feature to detect dementia. There have been several approaches to calculating the sentiment analysis of a sentence [74], among them three approaches based on lexicon are considered: TextBlob, sentiVader, and sentiWordNet. The approach based on sentiWordNet consists of calculating the average sentiment of all the content words, that remain after removing stop words, within a text or a speech sample. As shown in figure 12, only sentiBlob yielded a significant Anova score. Despite that, sentiment analysis features ranked high among the three other adopted feature selection approaches. For example, sentiBlob ranked among the top 25 information gain features, while both sentiWordNet and sentiVader ranked among the top 25 features of the three other feature selection approaches: reliefF, $\chi^2$, and information gain. These results confirm that dementia has a substantial impact on emotions and sentiment expression.

As shown in [27], there are several readability formulas that rely on different knowledge sources. Only the readability formulas that rely mainly on lexical characteristics are considered in this paper. Hence the following readability formulas have been selected: Gunning's Fog Score [75], Spache Score, and the Dale–Chall formula.

| Readability Formula | Equation | Variables |
|---|---|---|
| Gunning Fog Index | $FI = 0.4 \dfrac{N}{S} + 100 \dfrac{N_{complex}}{N}$ | N is the number of words<br>S the number of sentences<br>$N_{complex}$ the number of complex words, that have three or more syllables |
| Spache Score | $SS = (0.4121 \dfrac{N}{S}) - (-0.082 \dfrac{N_{unfam}}{N}) + 0.659$ | $N_{unfam}$ the number of complex words, that have three or more syllables |
| Dale Chall Score | $DCS = 0.1579 \left(\dfrac{N_{dif}}{N}\right) + 0.0496 \left(\dfrac{N}{S}\right)$ | $N_{dif}$ Number of difficult words, which are not in DC list of common words[10]. |

**Figure 13.** Equations of the three adopted Readability formulas

As seen in figure 12, the three adopted readability formulas yielded non-significant Anova results. Gunning Fog Index is among the top 25 features within the $\chi^2$ list. A probable reason for this outcome is that those formulas are not designed for spoken language.

Another way of using lexicon to detect dementia consists of building Ngram profiles of the two types of texts we have in the training data set: dementia and control. Three profiles are built for each of those two groups: bigram, trigram and fourgram. Then, during the feature extraction phase, three profiles are built for the current text and their distances are calculated with the corresponding training models (e.g. we calculated the distance between the bigram training model and the bigram model of the current text). Those distances are used as features and they are calculated as follows. For every Ngram in the text's model, the absolute value of the subtraction of the rank of this Ngram in the text from the rank of this Ngram in the level profile is added to a delta variable. The delta variable is the actual distance between the text and the level profile [76]. This approach gives us six distances: distFourgCt, distFourgDem, distTrigCt, distTrigDem, distBigCt, and distBigDem. As seen in figure 12, the six features related to the Ngram profiles yielded significant results. Despite the significance of the results of these features, it is worth noting that they are known to have issues with generalization when used outside of the specific task they are trained in.

---

[10] The Dale-Chall word list of common words: http://www.readabilityformulas.com/articles/dale-chall-readability-word-list.php

## 5. Evaluation and Discussion of Classification Evaluation and Results

The goal of this section is to provide a quantitative evaluation of the predictive power of the features presented in section 4. The following Machine-Learning algorithms have been used in the experiments: Logistic Regression (LR), Support Vector Machine (SVM), Adaptive Boosting (AB), bagging [77], and Random Forest (RF) [78] and eXtreme Gradient Boosting (XG) [79]. In addition, two versions of Mulilayer Perceptron were used. The first (MLP) has the following parameters: max_iter=200, hidden_layer_sizes= 50, activation function: tanh', solver=adam, alpha=1e-8, while the second version (NN) has the following parameters: max_iter=100, hidden_layer_sizes=40, activation function: relu, solver=adam, alpha=0.0001. These algorithms were selected for their better performance after having done some experiments with other algorithms, such as Decision Trees, and Naïve Bayes. All experiments were implemented in Python using Scikit-learn [80]. To measure the performance of the different algorithms, both accuracy and F1-score are reported when relevant, otherwise only F1 is reported (figure 14).

$$\text{Accuracy} = \frac{TN + TP}{N} \qquad \rho = \frac{TP}{TP + FN}$$

$$\pi = \frac{TP}{TP + FP} \qquad F1 = 2\frac{\rho * \pi}{\rho + \pi}$$

**Figure 14.** Where *N* is the number of samples, TP, FP and FN are the number of True Positives, False Positives, and False Negatives, respectively

To cover the different aspects of the built classifiers, three evaluation scenarios are adopted: comparing the types of features, selecting the optimal subsets of features based on the four adopted features selection methods, and identifying the optimal length of the input sample. In the next sections, the results of the evaluations are followed by a discussion.

### 5.1 Comparison of Lexical Types

In section 4, we have seen the individual impacts of lexical features, as they are presented by their types. Here we try to answer the question of significance of the features grouped by type in the classification of transcribed speech samples produced by patients with AD vs the ones produced by normal control subjects.

From a methodological point of view, although this study does not pretend to cover all the lexical features, it nonetheless collected a large number of lexical features per type. Hence, this comparison should give a general tendency. Before presenting the actual classification results per type, it is necessary to present the percentages of the features that made it to the top 50, according to the four adopted feature selection methods: Anova, reliefF, $\chi^2$, and information gain (figure 15).

As seen in figure 15, there are some disagreements between the different feature selection methods about the selected percentages of the lexicon areas. In all the feature selection methods, *diversity* has about half of the features, which is expected given the large number of features of this type. *Lexical sophistication* has between 6-10% of the features, while *lexical density* holds less than ten percent in all the feature selection approaches. The number of specificity features varies between 2% and 6%. *Psycholinguistics* has varying shares, between 6% and 16%. Finally, *miscellaneous* occupies about 20% in the four feature selection approaches.

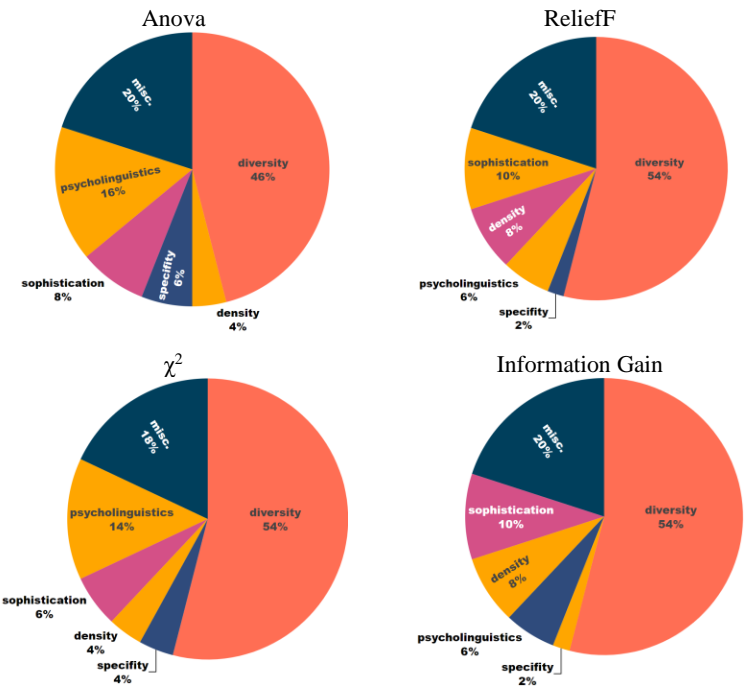

**Figure 15.** Rankings of the top 50 Features According to four different Feature selection Methods

The question now is whether these numbers of features have an impact on the accuracy and F1 scores of the classification with the different lexical types. Because of the special nature of the Ngrams of words, two models were built out of miscellaneous features: *misc1* and *misc2*. The model *misc1* includes the Ngrams profiles, while *misc2* includes all the miscellaneous features, except for the Ngrams profiles. The results of the classification are reported in figure 16, where the eight used machine learning algorithms are trained on the ADDReSS training data and tested on the testing data of the same data set.

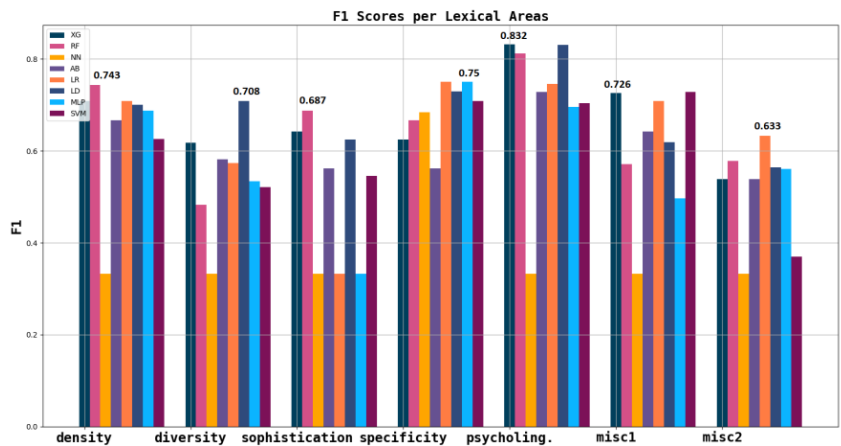

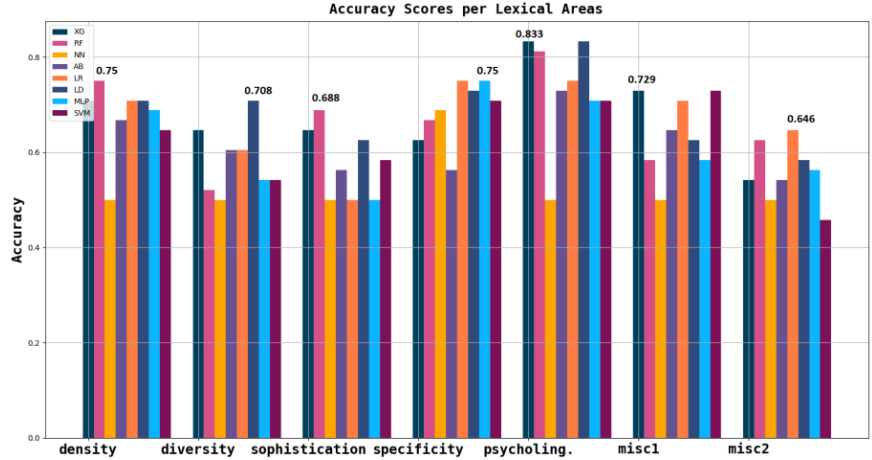

**Figure 16.** F1-scores and Accuracies of the lexical areas, on the ADDreSS Challenge data set

Here are some observations from figure 16. The *psycholinguistics* features provide the top classification results with the XG machine learning algorithm, with over 0.8 for both accuracy and F1. These results confirm the high impact of the psycholinguistic features, as demonstrated in table 1. The model *misc1* gives higher results than *misc2* because it includes distances between Ngrams profiles. The overall ranking of the features doesn't seem to systematically impact the results. For example, although *diversity* accounts for about half of the top 50 features and *density* and *specificity* account respectively for less than 10%, yet both density and specificity provide higher F1 and accuracy scores than diversity. Besides, all the lexical areas achieved lower than the best performance reported in the ADReSS Challenge, which is 0.89 for both F1 and Accuracy. This is probably due to the fact that large groups of features, such as the diversity features group, contain noisy and partially redundant features. Nonetheless, all the lexicon complexity types provide decent classification results, with *misc2* providing the lowest performance. No machine learning dominates with the different models. For example, XG provides the best performance with *psycholinguistics*, and *misc1*, RF provides the best performances with *density* and *sophistication*, MLP provides the best performance with *specificity*, SVM provides the best performance with *misc1*, and the best performance with *misc2* is provided with Logistic Regression. Finally, no substantial differences are observed between accuracy and F1 scores.

To better understand the results of the classifiers, an examination of the confusion matrices of the four top classifiers is provided in table 2.

| Psycholinguistics with XG | | | Density with RF | | |
|---|---|---|---|---|---|
| | Control | AD | | Control | AD |
| Control | 92% | 8% | Control | 92% | 8% |
| AD | 25% | 75% | AD | 41% | 59% |
| Misc1 with MLP | | | specificity with MLP | | |
| | Control | AD | | Control | AD |
| Control | 16% | 84% | Control | 75% | 25% |
| AD | 0% | 100% | AD | 0.25% | 0.75% |

**Table 2** Confusion matrices of the four leading classifiers, with the lexical areas: *psycholinguistics*, *density, misc1, and specificity*

As shown in table 2, both *psycholinguistics* with XG and *density* with RF do a better job at detecting control cases, while specificity of MLP's performance in detecting control and dementia cases is identical. On the other hand, *misc1* with MLP does a perfect job at detecting dementia and a poor job at detecting control. This suggests that the outcome pattern depends on the combination of the set of features and the machine learning algorithm.

It would be interesting to test how the different lexical areas scale out when applied to texts of different types. Hence, a scalability evaluation is conducted by training the classifiers on the ADReSS training data and testing them on the three data sets extracted from the novels of Iris Murdoch and Agatha Christie and from President Reagan's speeches (see section 3.3). Given that no substantial differences were observed between F1 and accuracy scores, only F1 results will be reported.

The results of the scalability test on the data set extracted from Iris Murdoch novels are presented in figure 17.

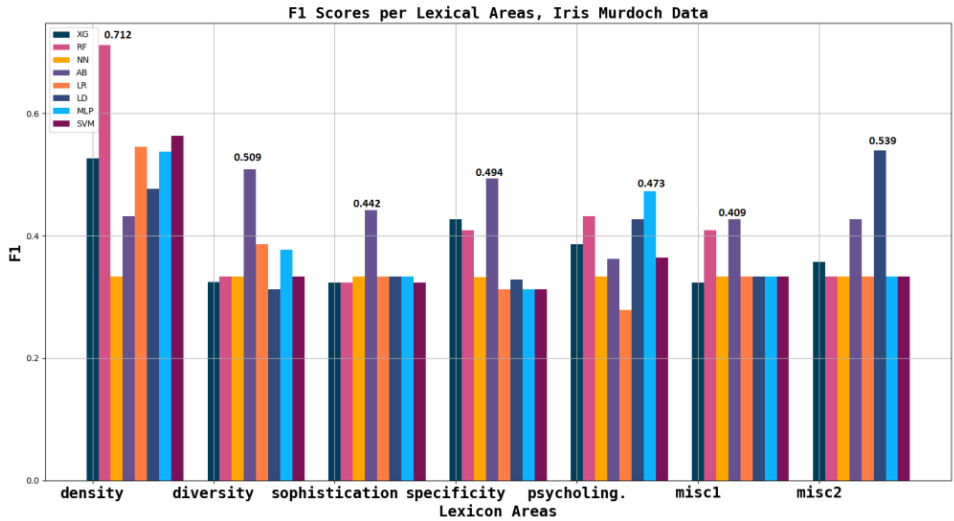

**Figure 17.** F1 scores of the Scalability evalution Conducted on Iris Murdoch data Set

As seen in figure 17, density with Random Forest achieves the highest F1 score, while diversity's best score is much lower. As expected, *misc2* did a better job at scaling up than *misc1,* which relies on Ngrams profiles.

The scalability results on the data of Agatha Christie are presented in figure 18.

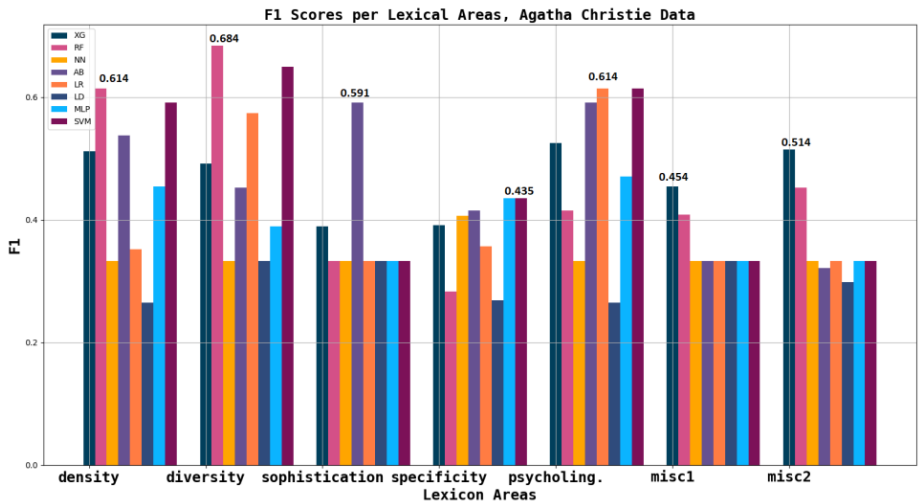

**Figure 18.** F1 scores of the scalability evaluation on Agatha Christie's data

The first observation about the results in figure 18 is that, unlike the F1 scores on Iris Murdoch's data, *diversity* scales better than *density* and that *sophistication* and *psycholinguistics* scale better here as well. The observation about the difference between *mis1* and *misc2* is confirmed here as well. This difference between the data of the two authors can be attributed to either the difference in their writing style, to the specific type of dementia they had, or a combination of both.

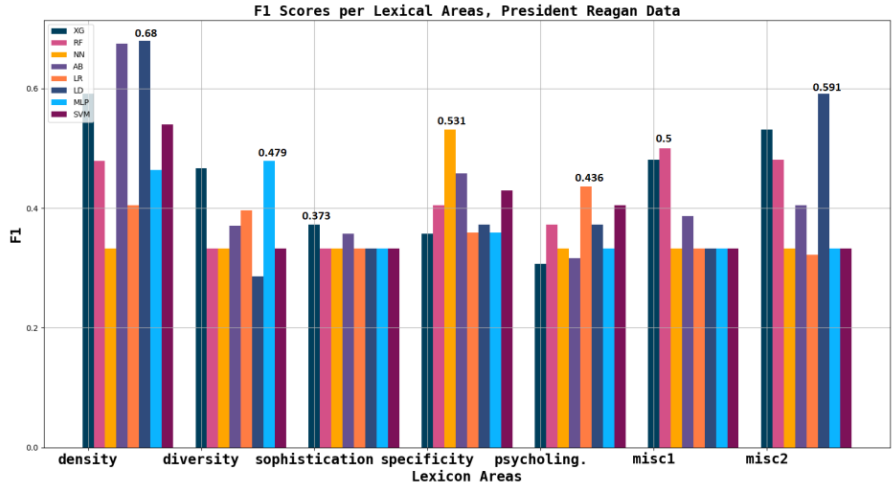

**Figure 19.** F1 scores of the scalability evaluation on president's Donald Reagan's data

The patterns of the results on President Reagan's data (figure 19) is closer to the one of Iris Murdoch, where *density* scales better than *diversity* and the rest of the lexical types. *Sophistication* and *psycholingusitics* don't do as well here as with Agatha Christie's data.

Let's examine some confusion matrices of the top classifiers to understand the generalization patterns.

| Density with XG, Ronald Reagan | | |
|---|---|---|
| | Control | Dementia |
| Control | 45% | 55% |
| Dementia | 25% | 75% |

| Diversity with RF, Agatha Christie | | | | Density with RF, Iris Murdoch | | |
|---|---|---|---|---|---|---|
| | Control | Dementia | | | Control | Dementia |
| Control | 100% | 0% | | Control | 50% | 50% |
| Dementia | 42% | 58% | | Dementia | 5% | 95% |

**Figure 20** Confusion matrices of the two leading classifiers with the lexical areas: *psycholinguistics* and *lexical density*

As seen in figure 20, the error patterns are different depending on the machine learning algorithm, the test data, and the set of features. For example, with Iris Murdoch, the classifier can deal with dementia texts 95% of the time, while it has issues dealing with control cases. With Ronald Regan's data, the XG algorithm with density, the classifier also does a better job at dealing with dementia cases than control cases, but the overall score is lower than with Iris Murdoch's data. Finally, with Agatha Christie's data, the classifier does a perfect job dealing with control data, but it makes a lot of mistakes with dementia cases.

### 5.2 Continuous Feature Selection

Having explored a large number of lexical features, the question that needs to be addressed is the following. Which combinations of those features lead to the optimal classification result? To answer this question, four key feature selection methods have been adopted in the experiments: Anova, $\chi^2$, information gain, and reliefF. Every one of these methods provides a different ranking of the features. Hence, starting from the top feature, per feature selection method, until the entire set of features, all possible subsets of features will be tested with seven of the machine learning algorithms that have been used in section 5.1, NN has been discarded because it gave low results systematically. The features selected did not include the Ngram profiles, because of their issues with generalization. In this evaluation, the ADDReSS Challenge scenario was replicated. In other words, all the machine learning models were trained on the training data provided in ADDReSS Challenge website and tested with the test data from the same website. As mentioned before, besides being balanced, this data set makes it possible to compare the results reported by the other works that used the same evaluation scenario. The F1 results of this evaluation are presented in figure 21.

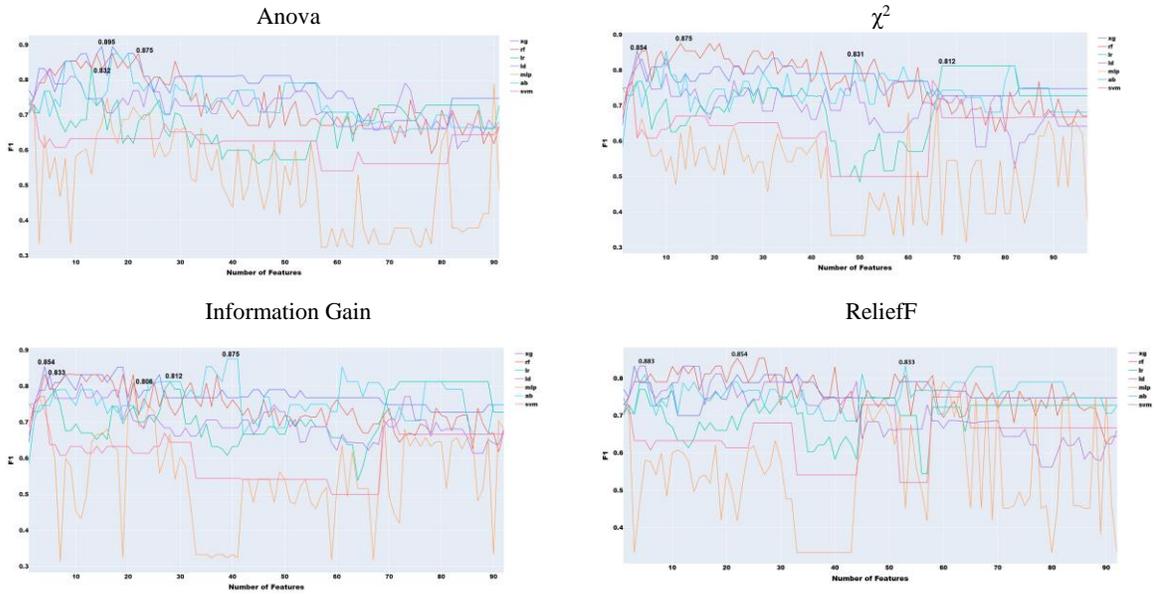

**Figure 21** F1 scores with seven machine learning algorithms and four feature selection methods: Anova, $\chi^2$, information gain, and reliefF

As shown in figure 21, quality matters more than quantity when it comes to machine learning models. AB is among the most demanding machine learning algorithms in terms of number of features, as it reached its top performances with $\chi^2$ and reliefF with more than 50 features. MLP is the least coherent, as its performance goes substantially up and down as the number of features increases. MLP is also the fastest to peak with $\chi^2$, as it requires only 3 features. However, it reached its peak with 90 features. The top overall result of 0.895 was achieved by Anova. The length of the top Anova list of features $A_{topAnova}$ has 15 features. This F1 score is equal to the best performance reported in the ADReSS Challenge. Consequently, this result shows that lexical information alone allows us to achieve such high performance. The top performance obtained with information gain is 0.875. It is achieved with 39 features, the length of $A_{topInfoGain}$.

To showcase the contribution of this paper, the union of the four top lists that gave the best performances, shown in figure 22, was performed to obtain the list of the top-most features in this study. This list is calculated as follows: $A_{topOverall} = A_{topAnova} \cup A_{topChi} \cup A_{topInfoGain} \cup A_{topReliefF}$. The length of top overall list $A_{topOverall}$ is 27 features. The distribution of those features is presented in figure 22.

| Type | List of Features |
|---|---|
| Newly Proposed | distLCH, rtHypo, distPath, rtNn, AVG_Glv50 |
| Newly Used | PMI, Kf_freq, Concreteness, gapVb, familiarity, Paivio, sentiVader, Brown, Colorado, rtIrregVB, sentiWordNet, VSM |
| Old | ANR, CTTR, NVR, GTTR, rtAdj, rtVb, rtAdv, discWrd, rtNouns, imageability |

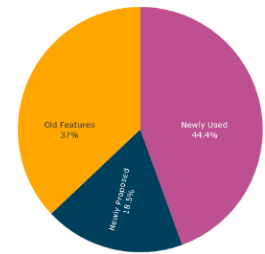

**Figure 22.** Distribution of the Most Impactful Features

As shown in figure 22, most of the features which made it to the top are either proposed in this paper or used for the first time for dementia detection. About a 37% of the features have been used in previous works.

### 5.3 Input Length Experiments

Another key question, that hasn't been addressed in previous works, to the limit of our knowledge, is the length of the input used in training and testing machine learning models. Knowing the optimal input size can help maximize the yield of machine learning algorithms. Besides, optimizing the input size helps design the psychological tests such that the right amount of data is gathered from the subjects, thus reducing the effort and cost of the process. Before we present the results, some basic statistics can help better understand the distribution and the disparities between the lengths of the speech samples within the ADReSS data set (table 3).

|  | Min | Max | Mean | SD |
|---|---|---|---|---|
| Training | 44 | 559 | 135.129 | 73.164 |
| Testing | 48 | 523 | 138.833 | 84.096 |

**Table 3** Stats of the numbers of tokens of speech samples within the training and testing parts of the ADDReSS data set

A test is conducted with gradual inputs within the range between 5 to 225 tokens and an increment of 1. The upper bound of 225 is chosen because it exceeds the mean length, by about one standard deviation, in both test and training data. This test is repeated with the four optimal numbers of features identified in the feature selection approaches. Given that the data in the chosen upper bound is higher than the minimum numbers of tokens gathered in some entries, the tests will be conducted with whole data gathered from a subject when it is lower than the chosen upper limit, and with a sub part of the data gathered when its length exceeds or is equal to the upper limit. The F1 results are presented in figure 23. Accuracy results were similar to the F1 scores; therefore, they are not presented.

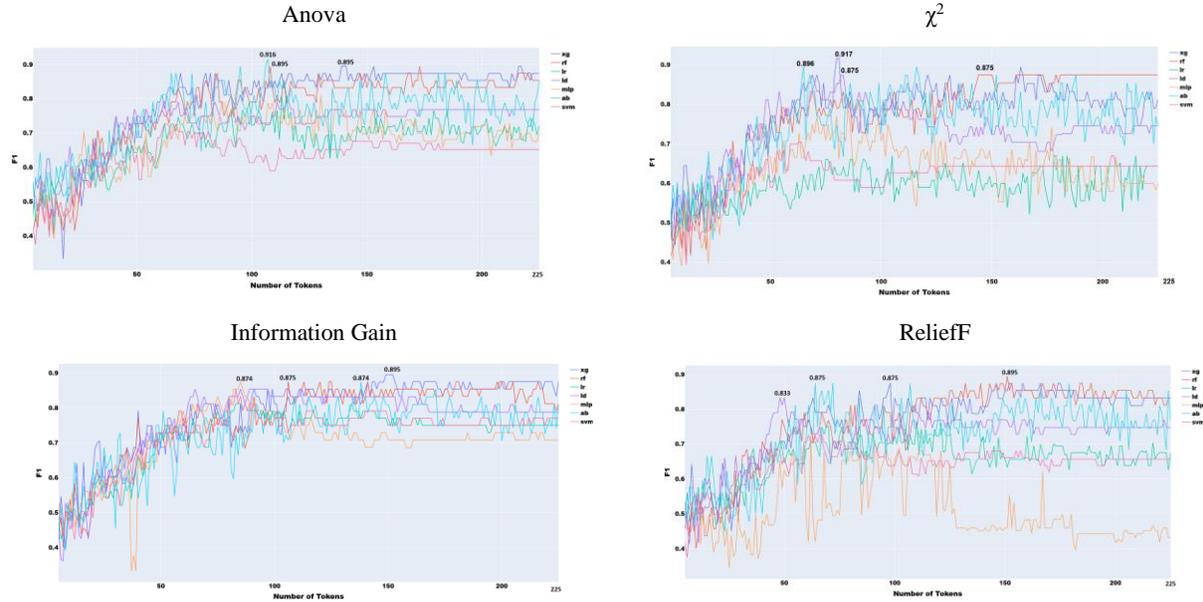

**Figure 23.** Experiments conducted with gradual lengths between 5 and 225 tokens, with an increment of 1

As we can see in figure 23, performances with the four feature selection methods have been slightly improved by reducing the input length. For example, XG peaked at 0.895, with around 150 tokens, with the optimal information Gain features. While the same algorithm peaked at 0.854 without input limitation. Similar improvements were observed with other machine learning algorithms, such as RF, and MLP, meaning that the benefit of limiting the input length is not proper to a specific machine learning algorithm. The best overall performance of 0.917 was obtained with the optimal set of $\chi^2$ features and XG, with an input limited to 80 tokens. This score slightly exceeds the best reported F1 score in the ADReSS Challenge.

To study the impact of input length on generalization to other tasks, similar tests were conducted on Iris Murdoch data. So, the best set of features according to the four adopted feature selection methods were used along with input lengths ranging from 5-225, with increments of 1 (figure 24).

As seen in figure 24, the best overall F1 performance of about 0.78 is achieved with AB, information gain, and input length of 12 tokens. This shows that, despite the decline of performance when training and testing are done on two different data sets, lexical models can still achieve a reasonable generalization to other tasks when limiting the input length. This result is very promising given the known differences between spoken language (used for training) and written samples used for testing [81].

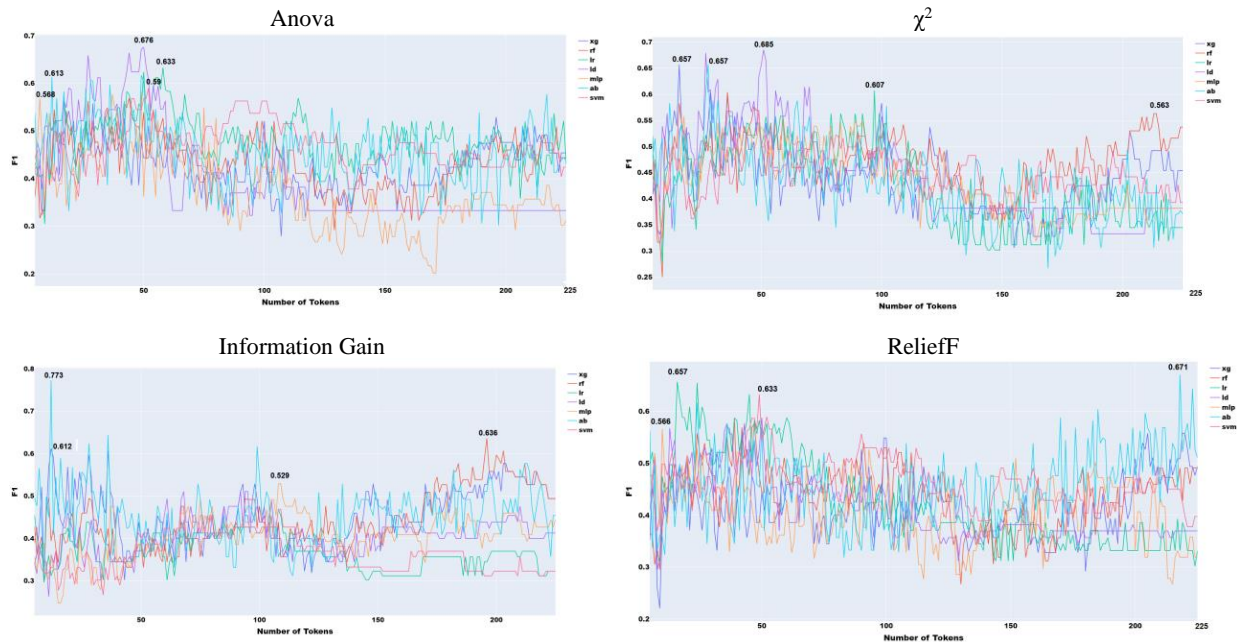

**Figure 24** Generalization test on Iris Murdoch data with gradual input sizes between 5 and 225 tokens

## 6. Conclusion

This paper is about exploring the impact of AD on the lexicon and how lexical features can help build an automatic classifier that can predict whether a transcribed speech sample or a written text is produced by someone with AD.

The set of 99 features that was covered in this study includes features that have been used before with AD studies, such as ANR, TTR and some of its derivatives like Brunet Index, and GTTR. The set of covered features also includes features that have been explored in other application areas, such as sentiment analysis, gap verbs, irregular verbs, and sentiment analysis. Finally, some new features have been proposed by the author, such as content focus, Wu-Palmer Similarity, Path similarity, and the ratio of hyponyms to the words. The individual examination of this wide body of features allowed to shed more light on the impact of dementia on the lexicon in general. Among the 27 features that helped reach the peak machine learning classification scores, about 63% were here either proposed or used for the first time to detect dementia. Among the findings that are worth mentioning, is the substantial impact of dementia on the usage of emotional lexicon as measured by lexicon-based sentiment analysis algorithms. It is also worth noting that the features that made it to the top list are distributed over all the lexical complexity areas. This distribution confirms that interest of conducting a study on a large body of features to get a set of complementary pieces of the puzzle.

The second part of this paper is about using the identified features in the first part to build machine learning models to detect dementia based on written or spoken language samples produced by patients with AD and normal control subjects. The experiments on features grouped by type of lexical complexity, showed that the *psycholinguistic* features were the most important when the test was performed on data of the same task. Depending on the data used, *density* and *diversity* scaled better than the other types of lexical complexity when tested on a different task. All the groups of features by lexical complexity types performed lower than the state of the art. Features selection experiments, with the four adopted feature selection approaches, helped optimize the performance of the seven adopted machine learning algorithms. With the 15 best features selected by Anova and the XG machine learning algorithm, a state-of-the-art performance was achieved. The experiments with gradual input length showed that the input length is a key factor that influences the results, whether when it is too short or too long. Hence, an improvement in the performance of all the best models was achieved when the test was conducted with an optimal input length. The reported performance of 0.917 was obtained with an input length of 81 tokens, and the optimal chi square set of features with the XG algorithm. This performance is a slight improvement compared to the best result reported in the ADReSS challenge. Furthermore, the scalability evaluation showed that optimizing the input length can lead to performance improvement as well.

Overall, this paper showed the importance of the lexicon as a means to detect AD. Nonetheless, it would be interesting in a future work to combine lexical features with other linguistic and acoustic models to further optimize the classification performance.

Conducting experiments with other feature selection approaches can help identify other optimal sets of features and perhaps bring further improvement to the classification performance.

**Limitations**

Unlike previous works, this study is not limited to a single data set. However, despite the variation in the used data set, the presented results are likely to improve with the increase in the number of data samples. This study is limited to the lexical factors, which have been shown to play a key role in dementia detection in several previous works. However, lexical factors are not the only ones that can help detect dementia, others like acoustic factors, morphological factors, and syntactic factors need to be explored. The Ngram profiles used in this study are, as we showed in the generalization tests, not good at scaling up to other tasks.